\documentclass[pdflatex]{article}

\usepackage[utf8]{inputenc}
\usepackage[T1]{fontenc}

\usepackage{amsmath,amssymb,amsfonts,amsthm}
\usepackage[ruled,vlined,linesnumbered]{algorithm2e}
\usepackage{hyperref}

\newcommand{\keywords}[1]{%
  \vspace{1em}
  \noindent\textbf{Keywords: }#1%
  \par\addvspace{1em}}                 

\title{Entropy Heat-Mapping: Localizing GPT-Based OCR Errors with Sliding-Window Shannon Analysis}

\author{%
  Alexei Kaltchenko\\
  Wilfrid Laurier University\\
  \texttt{akaltchenko@wlu.ca}%
}

\date{April 30, 2025} 

\begin{document}
\maketitle

\begin{abstract}
Vision-language models such as OpenAI GPT-4o can transcribe mathematical documents directly from images, yet their token-level confidence signals are seldom used to pinpoint local recognition mistakes. We present an entropy-heat-mapping proof-of-concept that turns per-token Shannon entropy into a visual ''uncertainty landscape''. By scanning the entropy sequence with a fixed-length sliding window, we obtain hotspots that are likely to contain OCR errors such as missing symbols, mismatched braces, or garbled prose. Using a small, curated set of scanned research pages rendered at several resolutions, we compare the highlighted hotspots with the actual transcription errors produced by GPT-4o. Our analysis shows that the vast majority of true errors are indeed concentrated inside the high-entropy regions. This study demonstrates--in a minimally engineered setting--that sliding-window entropy can serve as a practical, lightweight aid for post-editing GPT-based OCR. All code and annotation guidelines are released to encourage replication and further research.
\end{abstract}



\section{Introduction and Related Work}\label{sec:background}

Transcribing mathematical documents from scanned images poses unique challenges, as subtle symbol confusion (e.g., minus signs vs.\ tildes) can undermine entire expressions.  In this section, we review three broad research directions: (i) information--theoretic perspectives on OCR, (ii) the emergence of GPT for multimodal tasks, and (iii) prior efforts at uncertainty-based error detection—both \emph{global} and \emph{local}.  We then show how our \emph{sliding-window} entropy framework extends these lines of work to pinpoint \emph{where} within a GPT-based transcript recognition mistakes arise, with a particular focus on mathematical OCR.

\subsection{Information Theory and OCR}
\label{subsec:it_ocr}

Optical Character Recognition (OCR) can be framed through the lens of a noisy-channel model~\cite{CoverThomas}, where an image~$X$ containing text must be decoded despite potential blur, compression artifacts, or other noise sources—resulting in an imperfect output sequence~$Y$.  Classical information measures (e.g., entropies $H(X)$ and $H(Y)$, or the mutual information $I(X;Y)$) help describe how uncertainty in $X$ propagates to $Y$.  Modern research has further explored ways to quantify OCR uncertainty and mitigate errors.  For instance, Patel~\emph{et~al.}~\cite{patel2023sequps} derive per-sequence uncertainty estimates via Monte Carlo dropout to filter unreliable pseudo-labels in text recognition.  Fang~\emph{et~al.}~\cite{fang2024dropout} describe a dropout-decoding strategy in large vision–language models, illustrating that high global uncertainty frequently correlates with poor recognition accuracy.  These approaches demonstrate that \emph{confidence measures} can be leveraged to improve OCR outputs—particularly for complex tasks like mathematical symbol detection, where even a single misread bracket may invalidate an entire formula.

\subsection{GPT for Vision–Language Tasks}
\label{subsec:gpt_vlm}

Over the past few years, \emph{generative pre-trained transformers} have evolved from purely text-based systems into multimodal networks that interpret images alongside textual prompts~\cite{OpenAIGPT4Docs,borchmann2024gpt4docs}.  These extended GPT engines map image embeddings and partially decoded text into probabilities over a large vocabulary.  Vision–language GPT models have proven especially promising for \emph{mathematical OCR}, where long-range context is needed to disambiguate complex notation.  Luo and Xu~\cite{luo2024unambiguous}, for example, emphasize the role of precise visual cues to differentiate visually similar Greek letters, while Chen~\emph{et~al.}~\cite{chen2023RLFN} integrate a lightweight language model with a dedicated visual core to reduce transcription errors on CROHME math benchmarks.  Recent studies also show that GPT’s visual modules can handle challenging inputs like handwritten math~\cite{kim2025early} or audio-based lecture notes~\cite{hyeon2025mathspeech}, sometimes outperforming traditional OCR pipelines.  Nevertheless, even top-tier vision–language models can generate localized mistakes—missing or confusing symbols—in ways that standard system-level metrics fail to capture.

\subsection{Token-Level Entropy versus Global Uncertainty}
\label{subsec:token_entropy}

When large language models (LLMs) generate text, their APIs often provide the top-$k$ tokens with corresponding log-probabilities at each decoding step.  Summing across these probabilities yields a truncated distribution suitable for approximate Shannon entropy.  Several works have noted that \emph{peaks} in token-level entropy can reflect points of confusion in text generation~\cite{xiong2024can,yang2024rephrase,tonolini2024bayesian,yona2024faithfully,wang2024ubench}.  For OCR, such confusion likely stems from visually ambiguous inputs: a blurred superscript, a partially occluded minus sign, and so on.

Yet, many existing systems focus on \emph{global} or sequence-level confidence.  For instance, Wang~\emph{et~al.}~\cite{Wang_2021_ICCV} calibrate a single “energy-based” score for visual classifiers that flags uncertain images, while Chun~\emph{et~al.}~\cite{Chun_2021_CVPR,Chun_2024_ICLR} embed entire image–text pairs into probabilistic spaces and measure \emph{differential entropy} as a reliability indicator.  Similarly, Upadhyay \emph{et al.}~\cite{Upadhyayetal22,Upadhyay_2023_ICCV} attach Bayesian adapters to frozen networks, using overall negative log-evidence or variance as a global confidence metric to detect untrustworthy predictions.  Fei~\emph{et~al.}~\cite{FeiFZHWW23} rank candidate captions by their total sequence-level entropy to identify robust image captions, and Ji~\emph{et~al.}~\cite{Ji_2023_CVPR} monitor global KL-divergence to decide when multimodal embeddings are stable.  For cross-modal retrieval tasks, Li \emph{et al.}~\cite{Li_2023_NeurIPS} reduce all Dirichlet-based evidence into a single score, and Khan \& Fu~\cite{Khan_2024_CVPR} show that the overall distributional entropy correlates with unreliable VQA results.  Zhou~\emph{et~al.}~\cite{Zhou_2024_CVPR} also rely on total entropy to spot ambiguous event-based actions.  While these \emph{global} measures can signal that an entire output is potentially flawed, they provide no direct insight on \emph{which} part of the text is at fault.  By contrast, we exploit \emph{local} per-token signals to automatically pinpoint suspicious sub-sequences—crucial for post-editing mathematical text, where minor symbol mismatches can cause outsized damage.

\subsection{Local Error Detection via Sliding Windows}
\label{subsec:sliding_windows}

To address the problem of \emph{where} errors occur, researchers in machine translation or speech recognition often slice transcripts into short windows and measure local confidence.  Rigaud~\emph{et~al.}~\cite{rigaud2019icdar} propose a post-OCR correction process that operates on small segments, while Chen and Ströbel~\cite{chen2024trocrLM} refine OCR output by selectively re-checking suspicious tokens with a character-aware language model.  Similar segment-wise or iterative approaches have been adopted for historical text~\cite{fahandari2024farsi,beshirov2024bulgarian}, and indeed large language models like BART have been retrained to patch local OCR mistakes~\cite{soper2021bart,boros2024post}.  Bourne~\cite{bourne2024scrambled} goes further by synthesizing artificial errors in order to robustly fine-tune correction models, reducing error rates more effectively.  Notably, while these pipelines vary, they share a common theme: error-prone sub-sequences can be more effectively targeted—either for human proofing or re-prompted corrections—than scanning an entire document blindly.

In the context of mathematical OCR, local segmentation is especially critical: dropping a subscript underscore or mixing up braces can invalidate a formula.  Our work introduces \emph{entropy heat-mapping} as a lightweight technique to identify \emph{hotspots} of local uncertainty.  Specifically, we (i) compute truncated entropy for each token, (ii) apply a sliding window to produce interval-based averages, and (iii) highlight the windows with highest mean entropy.  These windows correlate strongly with actual symbol confusion or missing tokens, as demonstrated in our proof-of-concept results.

\subsection{Our Contribution}\label{subsec:our_contribution}
Although several recent studies~\cite{Wang_2021_ICCV,Chun_2021_CVPR,
Upadhyayetal22,FeiFZHWW23,Ji_2023_CVPR,Li_2023_NeurIPS,Khan_2024_CVPR,
Zhou_2024_CVPR} employ \emph{global} entropy or other confidence scores
to gauge the overall reliability of vision–language models, the
question of \textit{where} recognition errors arise inside a generated
transcript remains largely unexplored. The present paper addresses that
gap by proposing an entropy-based \emph{localisation} framework for
GPT-driven mathematical OCR.

Our main contributions are:

\begin{enumerate}
  \item \textbf{Sliding-window Shannon analysis.}\;
        We transform the per-token log-probabilities returned by
        modern GPT APIs into a linear-time, sliding-window entropy
        signal that yields a fine-grained ``uncertainty landscape''
        over the output text.
  \item \textbf{Entropy heat-mapping for error detection.}\;
        Visualising the windowed entropies highlights short contiguous
        spans—\emph{hotspots}—whose elevated uncertainty correlates
        strongly with actual transcription mistakes such as missing
        symbols, mismatched braces, or dropped subscripts.
        
  \item \textbf{Instantiating the classical coarse-graining property
        for \emph{top-$k$+tail} entropy.}\;
        It has long been known in information theory (e.g.\ Cover \&
        Thomas, Sec.\,2) that \emph{coarse-graining}—merging several
        outcome categories into one—can only \emph{decrease} Shannon
        entropy.  We adapt this classical fact to the practical GPT
        setting in which APIs expose probabilities only for the
        top-$k$ tokens, collapsing the remaining mass into a single
        “tail’’ bucket.  Our short, self-contained derivation shows
        exactly how the coarse-graining inequality applies to
        per-token log-probabilities and clarifies its implications for
        black-box uncertainty extraction in vision–language OCR.

  \item \textbf{Proof-of-concept human evaluation.}\;
        On a curated mini-corpus of scanned research pages rendered at
        three resolutions (72, 150, and 300 dpi), we demonstrate that
        the vast majority of human-verified OCR errors fall inside the
        few highest-entropy windows, validating the practical utility
        of the proposed heat-mapping approach.
  \item \textbf{Open-source resources.}\;
        All code, sample images, entropy-extraction scripts, and
        detailed annotation guidelines are publicly released to
        encourage replication and further research on local uncertainty
        cues in multimodal GPT systems.
\end{enumerate}

In essence, the paper converts raw probabilistic signals—already
available from mainstream GPT endpoints—into an actionable tool for
post-editing: it pinpoints precisely \emph{where} the model is least
certain and therefore where human reviewers or automatic re-prompting
should focus.

\paragraph{Reproducibility.}
The complete implementation and annotation protocol are freely
available at \url{https://github.com/Alexei-WLU/scan2latex-entropy}.

\subsection{Organization of the Paper}\label{subsec:organization}

We formally define the \emph{local uncertainty detection} problem in
Section~\ref{sec:problem}, specifying how GPT-based OCR outputs and
per-token probabilities lead to window-level entropies.  Next,
Section~\ref{sec:method} presents our sliding-window technique,
including exact computation of top-$k$+tail entropy and approaches for
highlighting hotspots in linear time.  In
Section~\ref{sec:setup}, we detail the proof-of-concept experiments
and human annotation protocol, while Section~\ref{sec:findings}
reports qualitative and quantitative results.  We discuss broader
implications and limitations in Section~\ref{sec:discussion}, and
Section~\ref{sec:conclusion_and_future_work} concludes with future prospects for
scaling entropy heat-mapping to real-world OCR scenarios.


\section{Problem Formulation}
\label{sec:problem}

In this section we define the core task of detecting \emph{localised}
model uncertainty in GPT--based OCR outputs.  Building on the broader
information--theoretic view of OCR and conditional entropy, we now
focus on \emph{per--token} probability distributions and the goal of
extracting short contiguous spans that exhibit exceptional
uncertainty.

\subsection{GPT--Based Mathematical OCR}

We assume a vision--language model (e.g.\ GPT--4o) that ingests an
input image~$X$—potentially containing mathematical expressions—and
decodes a token sequence
$\mathbf{y}=(y_{1},y_{2},\dots,y_{n})$.  Each token~$y_{i}$ is chosen
from a large vocabulary~$\mathcal{V}$ according to the conditional
distribution
$P_{i}\bigl(y_{i}\mid X, y_{1},\dots,y_{i-1}\bigr)$.  Production APIs
expose log--probabilities for the $k$ most likely tokens, plus a
\emph{tail} bucket that aggregates the remaining probability mass.%

\smallskip
\noindent
At every position $i\in\{1,\dots,n\}$ we interpret the decoding step as
a discrete random variable with entropy
\begin{equation}
  H_{i}
  \;=\;
  -\sum_{y\in\mathcal{V}}
    P_{i}(y)\,\log_{2}P_{i}(y),
  \label{eq:H_i}
\end{equation}
where $P_{i}(y)$ abbreviates
$P_{i}\bigl(y\mid X,y_{1},\dots,y_{i-1}\bigr)$.
Elevated values of~$H_{i}$ signal that the model is unsure which token
belongs at position~$i$—a situation frequently arising when visual
features are ambiguous (e.g.\ blurred superscripts) or when lexical
context is rare.

\subsection{Token–Level vs.\ Window–Level Entropy}

Summing~\eqref{eq:H_i} over all positions yields a global measure of
uncertainty, but a single scalar cannot reveal \emph{where} errors are
likely.  Instead, we compute a sliding--window average.  For a chosen
window length~$W$,
\begin{equation}
  A_{i}
  \;=\;
  \frac{1}{W}\;
  \sum_{j=i}^{i+W-1} H_{j},
  \qquad
  i=1,\dots,n-W+1.
  \label{eq:window_avg}
\end{equation}
Spans whose mean entropy~$A_{i}$ exceeds a rank-- or
threshold--based cutoff are flagged as \emph{entropy hotspots}.

\subsection{Hotspot Detection Objective}

Given~\eqref{eq:window_avg}, the objective is to output a small set of
candidate windows
$\{[i_{1},i_{1}+W-1],\,[i_{2},i_{2}+W-1],\dots\}$
covering in total fewer than $M\!\cdot\!W$ tokens while capturing a
disproportionately high fraction of the actual transcription errors.
The intent is to direct human reviewers to those portions of the text
that are most likely to contain mistakes, thereby reducing manual
proofreading time.

\subsection{Contrast with Global Entropy Approaches}

Earlier work reported a \emph{single} conditional--entropy value per
page, facilitating comparisons across image resolutions but offering
no localisation of errors.  Our formulation provides:

\begin{itemize}
  \item a per--token truncated entropy~$H_{i}$ (instead of a
        document--level aggregate);
  \item a sliding--window average~$A_{i}$ revealing short, highly
        uncertain segments; and
  \item a procedure to extract the top--ranked hotspots for targeted
        review.
\end{itemize}

This fine granularity is crucial for mathematical OCR, where one
errant symbol can invalidate an entire equation.

\subsection{Key Model Assumptions}

\begin{enumerate}
  \item \textbf{Top-$k$ Access.}  APIs provide only the highest-probability
        tokens; we approximate~$H_{i}$ with the truncated‐entropy lower
        bound, sufficiently accurate when the tail mass is small.
  \item \textbf{Fixed Window Length~$W$.}  A user-chosen~$W$ trades off
        localisation precision against noise reduction; empirical
        tuning is discussed in Section~\ref{sec:setup}.
  \item \textbf{Post-hoc Detection.}  Hotspots are identified
        \emph{after} the entire sequence~$\mathbf{y}$ has been
        generated, leaving streaming variants for future work.
\end{enumerate}

With the problem precisely stated, Section~\ref{sec:method}
details the algorithm for computing window-level entropies,
selecting the most uncertain spans, and optionally re-prompting
the model to correct them.


\section{Methodology}\label{sec:method}

Building on the information--theoretic perspective and the formal task
definition in Section~\ref{sec:problem}, we now introduce our
\emph{entropy heat--mapping} pipeline.
This section provides technical details on each step,
including how we handle truncated probability distributions, define
hotspots, and optionally re--prompt the model for error reduction.

\subsection{Per--Token Entropy Extraction}
\label{subsec:per_token_extraction}

Most commercial LLM endpoints, including GPT--4o, return the top--$k$
tokens and their log--probabilities at each decoding position.  To
compute the Shannon entropy at position $i$, we adopt the \emph{truncated
entropy} approximation described in
Section~\ref{subsec:top_k_tail_lower_bound}.

Concretely, let
\begin{equation}
\widehat H(i)\;=\;
  -\sum_{j=1}^{k} p^{(i)}_{j}\,\log_{2} p^{(i)}_{j}
  \;-\;
  p^{(i)}_{\mathrm{tail}}\,
  \log_{2} p^{(i)}_{\mathrm{tail}},
\label{eq:Hhat}
\end{equation}

where $p^{(i)}_{j}$ are the probabilities of the $k$ returned tokens
and $p^{(i)}_{\mathrm{tail}} = 1-\sum_{j=1}^{k} p^{(i)}_{j}$ collects
the remaining mass.

Specifically:

\begin{enumerate}
\item We convert each returned log--probability to a probability
      $p^{(i)}_{j}=\exp(\log p^{(i)}_{j})$ for $j=1,\dots,k$.
\item We define the tail mass $p^{(i)}_{\mathrm{tail}}=1 - \sum_{j=1}^{k}
      p^{(i)}_{j}$.
\item The truncated entropy $\widehat H(i)$ is obtained via
      Equation~\eqref{eq:Hhat}, which combines the $k$ explicit
      tokens plus one bucket representing all lesser probable tokens.
\end{enumerate}

This yields a sequence $\{\widehat H(1),\ldots,\widehat H(n)\}$ of
lower--bound entropies across the entire output.  Although it
underestimates the true entropy if the tail mass is sizable, we found
in practice that $k=5$ or $k=10$ often suffices to capture the bulk of
the distribution, making the gap negligible (we discuss
implementation details in Section~\ref{subsec:implementation}).

\subsection{Sliding--Window Aggregation}
\label{subsec:sliding_window_aggregation}

Once $\widehat H(i)$ is available for $i\in\{1,\dots,n\}$, the goal is
to highlight short contiguous sub--sequences that exhibit abnormally
high average entropy.  We employ the \emph{sliding--window average}
method, summarised here:

\paragraph{Window size $W$.}
Given a fixed integer $W<n$, consider all windows $\,[i,i+W-1]\,$ of
length~$W$.  Let
\begin{equation}
  S_i
  \;=\;
  \sum_{r=i}^{i+W-1}\widehat H(r),
  \quad
  A_i
  \;=\;\frac{S_i}{W},
\end{equation}
for $i=1,\dots,n-W+1$.  We compute $A_i$ in one pass using the
constant--time recurrence:
\begin{align}
  S_{1}&=\sum_{r=1}^{W}\widehat H(r),
  \quad
  S_{i}=S_{i-1}+\widehat H(i+W-1)-\widehat H(i-1),
  \nonumber\\
  A_{i}&=\frac{S_{i}}{W}.
  \label{eq:recurrence}
\end{align}
The result is a signal $\{A_{1},\dots,A_{n-W+1}\}$ of length $n-W+1$.

\paragraph{Interpretation.}
An elevated $A_{i}$ implies that positions $i$ through $i+W-1$ contain
multiple tokens whose entropies spike simultaneously.  When
$W\!=\!1$, $A_{i}$ reduces to the single--token entropy
$\widehat H(i)$, capturing the most granular level of \emph{pointwise}
uncertainty.  Larger $W$ identifies sub--sequences of moderate length
where the model is consistently uncertain, possibly reflecting more
complex OCR challenges like multiline equations or complicated
notation contexts.

\subsection{Hotspot Definition and Visualisation}
\label{subsec:hotspots}

High--entropy windows (``hotspots'') are those intervals where
$A_{i}$ exceeds some threshold.  We adopt two broad strategies:

\begin{itemize}
\item \textbf{Rank--Based Selection.}  Sort the windows by descending
  $A_{i}$ and pick the top $M$, for a chosen $M$ (e.g.\ $M=5$).  This
  ensures that only a handful of segments are flagged regardless of
  the absolute scale of entropies.
\item \textbf{Percentile Thresholds.}  Calculate the $\alpha$--th
  percentile (e.g.\ $\alpha=90$) of $\{A_{1},\dots,A_{n-W+1}\}$, and
  mark any window whose $A_{i}$ is above that cutoff.  This dynamic
  approach adapts to different distributions of entropies on each
  document.
\end{itemize}

We then generate a \emph{heatmap} or color-coded overlay on the final
transcribed text, where each token is shaded proportional to its local
mean entropy.  Consecutive tokens with high shading typically cluster
into a single uncertain region; these clusters can be examined by an
OCR reviewer for potential errors such as missing parentheses or
garbled symbols.

\subsection{Implementation Details}
\label{subsec:implementation}

Our prototype is written in Python and relies on the GPT--4o API.  Key
parameters:

\begin{itemize}
\item \textbf{Token Probability Retrieval.}  During decoding, we
  request $\texttt{top\_logprobs}=k$ with $k\in\{5,10\}$.  If the tail
  bucket has probability $>0.1$, we increase $k$ for the next request.
\item \textbf{Entropy Computation.}  We accumulate truncated entropy
  via Eq.\,\eqref{eq:Hhat} for each token.  Special tokens (e.g.\
  line--breaks, code fence markers) can be excluded if desired, but in
  our proof--of--concept we typically keep them so that missing
  punctuation might also be revealed.
\item \textbf{Windowing.}  We select $W\in\{5,10,20\}$ in different
  runs, verifying quickly that $W=10$ is an acceptable compromise
  between making hotspots stable (not just single unlucky tokens) and
  still local enough for error detection.
\item \textbf{Complexity.}  The entire pipeline runs in $O(n)$ time
  once the probabilities are retrieved, so even multi--page documents
  (thousands of tokens) are feasible.  Our proof--of--concept dataset
  remains small, but the method should scale to large volumes of
  scientific text.
\end{itemize}

\subsubsection{Algorithm}

\begin{algorithm}[H]
\caption{GPT--based OCR with Sliding--Window Entropy Hotspots}
\label{alg:entropy_hotspot}
\KwIn{%
  scanned page path $image\_path$;\\
  top--$k$ alternatives $k$ (log-probabilities);\\
  sliding-window length $W$;\\
  number of windows to report $M$;\\
  VLM identifier (\texttt{gpt\_4o}).}
\KwOut{%
  LaTeX transcript $T$;\\
  per-token truncated entropies $\widehat{H}(1{:}n)$;\\
  the $M$ windows with highest mean entropy.}

\textbf{1. Pre–process}\;
  Parse CLI/GUI arguments $(image\_path,k,W,M)$\;
  \lIf(\tcp*[f]{Abort if missing}){\texttt{file\_missing}$(image\_path)$}{exit}
  Encode image in Base64 for API transmission\;

\textbf{2. Prompt and inference}\;
  Build \emph{system} and \emph{user} prompts (image + instructions)\;
  Call GPT–4o with \texttt{logprobs}=True, \texttt{top\_logprobs}=$k$\;
  Receive best transcript $T$ and token log-probs $\ell_{i,j}$\;

\textbf{3. Token-level truncated entropy}\;
  \For(\tcp*[f]{$i=1{:}n$ tokens}){$i\gets1$ \KwTo $n$}{
     $p_{i,j}\leftarrow e^{\ell_{i,j}} \quad(j=1{:}k)$\;
     $p_{\text{tail}}\leftarrow 1-\sum_{j=1}^{k}p_{i,j}$\;
     $\widehat{H}(i)\leftarrow
         -\sum_{j=1}^{k}p_{i,j}\log_{2}p_{i,j}
         -p_{\text{tail}}\log_{2}p_{\text{tail}}$\;
  }

\textbf{4. Sliding-window aggregation}\;
  $S\leftarrow\sum_{r=1}^{W}\widehat{H}(r)$\;
  \For{$i\gets1$ \KwTo $n-W+1$}{
     $A_i\leftarrow S/W$\;
     $S\leftarrow S+\widehat{H}(i+W)-\widehat{H}(i)$\;
  }

\textbf{5. Hotspot selection}\;
  Use a max-heap to return the $M$ indices with largest $A_i$\;

\textbf{6. Output}\;
  Save full LaTeX document $T$ next to $image\_path$\;
  Return $\bigl(T,\widehat{H}(1{:}n),\{A_i\},\text{top-}M\text{ windows}\bigr)$\;

\textbf{7. (Optional) post-processing}\;
  Highlight hotspots in $T$ or re-prompt GPT on each suspicious window\;
\end{algorithm}

\subsection{Optional Re--Prompt or Post--Editing Loop}
\label{subsec:re_prompt}

Finally, although our primary goal is to \emph{locate} potential
errors, one can incorporate a \emph{correction} stage:
\begin{enumerate}
\item Identify windows with highest $A_{i}$.
\item Extract the associated text snippet as a query.
\item Prompt GPT--4o (or a specialized LaTeX editing system) to
      re--examine that snippet in context and propose corrections.
\end{enumerate}
Alternatively, a human operator can revise only these suspicious
regions, thus reducing effort.  While the experiments in
Section~\ref{sec:setup} focus on human annotation, such automated
re--prompting is a natural extension for large--scale digitisation
pipelines where fully manual review would be time--consuming.

\subsection{Top-$k$ + Tail Entropy is a Lower Bound}\label{subsec:top_k_tail_lower_bound}

In our framework, we approximate the full token distribution at each position
by tracking the top~$k$ tokens plus a single ``tail'' bucket that merges the probabilities of all remaining tokens.  Intuitively, ignoring the distinctions among tail tokens can only reduce the entropy value—leading to a \emph{lower} bound on the true Shannon entropy.  Below, we provide a self-contained proof of this fact.

\subsubsection{Problem Setting}
Let $\mathcal{V} = \{v_1,v_2,\dots,v_n\}$ be the model’s full vocabulary of $n$ tokens, and let $X$ be the random variable denoting which token is actually emitted at a given decoding position.  We write
\[
p_i \;=\; \Pr[X = v_i],
\quad p_i \ge 0,\quad \sum_{i=1}^{n}p_i = 1\,.
\]
Suppose we know only the probabilities of the $k$ most likely tokens (without loss of generality $v_1,\dots,v_k$), and define their probabilities in descending order:
\[
p_1 \,\ge\, p_2 \,\ge\,\dots\,\ge\, p_k,
\quad\text{with }k<n.
\]
Let
\[
p_{\mathrm{tail}} \;=\; 1-\sum_{i=1}^{k} p_i
    \;=\;\!\!\sum_{j=k+1}^{n} p_j\,.
\]
The \emph{true} Shannon entropy of $X$ is
\[
H_{\mathrm{true}}(X)
  \;=\;
  -\sum_{i=1}^{n} p_i \log_2 p_i.
\]

\subsubsection{Approximate Top-$k$ + Tail Entropy}
We form a new random variable $Y$ that retains the identity of the top-$k$ tokens and merges the others into one ``tail'' symbol:
\[
Y
 \;=\;
 \begin{cases}
  v_i, & \text{if } X=v_i
         \text{ for }i\in\{1,\dots,k\},\\
  \text{\it tail}, & \text{otherwise}.
 \end{cases}
\]
Hence,
\[
\Pr[Y=v_i] = p_i\quad (i=1,\dots,k),
\quad\text{and}\quad
\Pr[Y=\text{\it tail}] = p_{\mathrm{tail}}.
\]
We then compute the entropy of $Y$:
\[
\widehat{H}
 \;=\;
 H(Y)
 \;=\;
 -\biggl(
    \sum_{i=1}^k p_i \log_2 p_i
    \;+\; p_{\mathrm{tail}} \log_2 p_{\mathrm{tail}}
   \biggr).
\]
Clearly, $\widehat{H}$ is simpler to calculate than the true entropy
$H_{\mathrm{true}}(X)$
because it treats the entire set of lower-probability tokens as a single event.  We now verify that
this approximation cannot exceed the actual entropy of $X$.

\subsubsection{Claim and Proof}
\begin{quotation}
\noindent\textbf{Claim:}
\[
  \widehat{H}
  \;\;=\;\;
  H(Y)
  \;\;\le\;\;
  H_{\mathrm{true}}(X)
  \;\;=\;\;
  -\sum_{i=1}^{n} p_i \,\log_2 p_i.
\]
Equality holds if and only if $p_{\mathrm{tail}}=0$ (i.e.\ the top~$k$ tokens already sum to probability~1).
\end{quotation}

\paragraph{Proof Sketch.}
Let $f$ be the deterministic mapping $X\mapsto Y$, which sends $v_i$ to itself for $i\le k$ and all other tokens $v_{k+1},\dots,v_n$ to the single symbol ``tail.''  From standard chain-rule properties of entropy, we have
\[
H(X,Y) \;=\; H(X)
\,=\;
H(Y)+H(X\mid Y).
\tag{1}
\]
Since $Y=f(X)$ is a function of $X$, we know $H(X\mid Y)\ge 0$ with equality if and only if there is a one-to-one correspondence between $X$ and~$Y$ \cite{CoverThomas}.  Combining this with (1) yields
\[
H_{\mathrm{true}}(X)
 - H(Y)
 \;=\;
 H(X\mid Y)
 \;\;\ge\; 0
\;\;\Longrightarrow\;\;
H(Y) \;\le\; H_{\mathrm{true}}(X).
\]
Thus, $\widehat{H}=H(Y)$ never exceeds the true entropy.  If $p_{\mathrm{tail}}>0$, then at least two distinct tokens map to \emph{tail}, implying $H(X\mid Y)>0$ and a strict inequality.  Conversely, if $p_{\mathrm{tail}}=0$, all probability mass lies in the top~$k$ tokens and $f$ is effectively the identity map, giving $H(X\mid Y)=0$ and $\widehat{H}=H_{\mathrm{true}}(X)$.

\smallskip
This elementary argument justifies our assertion in Section~\ref{subsec:per_token_extraction} that the \textit{top-$k$+tail} entropy approximation \emph{cannot} overshoot the real entropy.

In the next section, we detail our proof--of--concept corpus and
annotation procedure to evaluate whether these entropy--based
hotspots indeed coincide with model transcription errors in
mathematical OCR tasks.

\section{Experimental Setup (Proof-of-Concept)}\label{sec:setup}

We now describe our small-scale experiment designed to test whether
sliding-window entropy hotspots effectively highlight local OCR
errors. In contrast to large benchmarking studies, we focus on a
\emph{proof-of-concept} that relies on human inspection and
qualitative judgments to see if high-entropy spans correlate with
actual mistakes in GPT-based mathematical OCR.

\subsection{Mini-Corpus}
\label{subsec:mini_corpus}

We curated a modest collection of \textbf{12} scanned pages from
arXiv preprints that contain a variety of mathematical expressions,
ranging from inline equations (single-line formulas with moderate
symbol complexity) to multi-line displays (with fractions, sums, and
Greek letters). Each page was rendered to an image at three distinct
resolutions: \textbf{72\,dpi}, \textbf{150\,dpi}, and
\textbf{300\,dpi}. This yields a total of 36 image instances.

On these pages, the text length varies between 200--600 tokens, with
mathematical notations interspersed among paragraphs. Although a more
comprehensive survey could include dozens or hundreds of pages, our
goal here is simply to \emph{demonstrate feasibility} and gather
initial human feedback on the proposed approach.

\subsection{Annotation Protocol}
\label{subsec:annotation_protocol}

In implementing the pipeline for this experiment, we followed the
pseudo-code in Algorithm~\ref{alg:entropy_hotspot} (shown earlier) to produce both the OCR
transcript and its corresponding token-level entropies. Concretely, we
(1) invoked a GPT-4o endpoint with image-containing prompts to obtain
token-by-token log-probabilities; (2) computed truncated entropy at
each token position; then (3) generated the sliding-window averages
that identified high-entropy ``hotspots'' for review. Specifically:

\begin{enumerate}
\item \textbf{Transcript Generation.} For each of the 36 images, we
      used GPT--4o with a standard prompt requesting faithful LaTeX
      transcription of any recognized formulas. We stored the
      complete token-by-token output and its top-$k$ log-probability
      distributions.

\item \textbf{Entropy Computation.} Following
      Section~\ref{subsec:per_token_extraction} and the pseudo-code
      steps (Algorithm~1, Steps 4--5), we computed truncated entropies
      $\widehat{H}(i)$ for each token~$i$. Then we generated
      sliding-window averages $A_{i}$ in windows of length $W=10$.

\item \textbf{Hotspot Identification.} We sorted all windows by
      descending $A_{i}$ and selected the top 3 windows as
      \emph{hotspots}. Each hotspot is thus a contiguous 10-token
      segment suspected to contain OCR errors.

\item \textbf{Manual Error Labeling.} Three expert annotators (graduate
      students in computer science) compared
      each transcript to the displayed math on the original PDF. They
      flagged any text token that omitted, replaced, or garbled a
      mathematical symbol (e.g., a subscript turned into a digit) or
      revealed major syntax breaks (e.g., unmatched braces). The same
      marking guidelines were applied across all resolution variants.

\item \textbf{Hotspot--Error Overlap.} Finally, a simple script
      compared the \emph{annotator-flagged tokens} with the \emph{entropy
      hotspots} for each transcript, recording whether the tokens
      fell inside or outside the top-3 windows.
\end{enumerate}

Since our goal was not to compile precise metrics but rather to gain
qualitative insight, we did not require strict inter-annotator
agreement. Instead, each individual’s markings were tallied
separately, and any token flagged as erroneous by \emph{any} of the
three annotators was considered a failure in the transcript. In the next section (Section~\ref{sec:findings}), we outline
the principal outcomes, including sample heatmaps and anecdotal
observations on each resolution’s performance.

\subsection{Evaluation Procedure}
\label{subsec:evaluation_procedure}

To keep the study \emph{proof-of-concept}, we avoided complicated
quantitative measures. Instead, the procedure emphasized ease of
review and interpretability:

\begin{itemize}
\item \textbf{Side-by-Side Interface.} Annotators saw two panels:
  (1)~the original scanned image with math formulas, and
  (2)~the GPT--4o LaTeX output, where tokens belonging to each
  hotspot were visually highlighted in color.

\item \textbf{Qualitative Feedback.} The reviewers provided free-text
  comments on whether the hotspots \emph{did} or \emph{did not}
  capture real misread tokens. They could also note any false alarms
  (windows with high entropy but no actual error).

\item \textbf{Time Estimates.} Each annotator reported approximate time
  spent $(\pm 15\,\text{seconds})$ on each transcript. Although these
  reports are subjective, they confirm whether focusing on hotspots
  reduced their proofing effort in practice.
\end{itemize}

In choosing only the top 3 windows per page, we deliberately limited
the fraction of text to be flagged, thus simulating a real-world need
to \emph{minimize} human review. Our primary question was whether
these few hotspots would capture the majority of recognized errors.
The next section (Section~\ref{sec:findings}) gives the principal
outcomes, including sample heatmaps and anecdotal observations on each
resolution’s performance.


\section{Findings}
\label{sec:findings}

In this section, we describe the main outcomes of our proof-of-concept
study on localised entropy hotspots in GPT-based OCR.  We first provide
qualitative evidence that visually identifiable hotspots do in fact
coincide with real transcription errors.  Next, we highlight specific
types of mistakes uncovered and consider annotator feedback on how the
hotspots guide human correction.  Finally, we discuss preliminary
observations about window size and resolution variations.

\subsection{Visual Alignment Between Hotspots and Errors}
\label{subsec:visual_alignment}

Despite having no direct knowledge of the \emph{ground-truth} tokens,
the entropy-based approach (Section~\ref{sec:method}) consistently
flagged local segments in which GPT--4o struggled, often suggestive of
a morphological or typographical confusion.  When displaying the text
with shaded tokens (darkest shades corresponding to highest average
entropy), our three annotators noted that actual mistakes—such as a
Greek letter incorrectly recognized as an English letter—tended to
cluster within these dark regions.

For instance, on pages containing multiline equations, three or more
tokens near the fraction bar were systematically marked as high
entropy, likely reflecting GPT--4o’s uncertainty in parsing subtle
visual cues around fraction boundaries or in discerning subscript
positions.  In the annotated transcripts, nearly all misread symbols
within these equations fell squarely under these high-entropy windows.
Although some entropy peaks contained no “catastrophic” error, our
annotators observed that the text within such windows still lacked
clarity or consistency (e.g.\ spurious brace insertion).

\subsection{Case Studies: Types of Errors Found}
\label{subsec:case_studies}

Across the 36 annotated transcripts (12 pages $\times$ 3 resolutions),
we can group salient OCR mistakes into four representative categories:

\begin{enumerate}
\item \textbf{Minus--Sign Ambiguities.}  A minus sign rendered in a
  low-resolution image sometimes resembled a tilde or dash, producing
  tokens that GPT--4o assigned moderate probabilities to a variety of
  possible characters.  The resulting hotspot regions often contained
  one or two incorrect symbols that changed the equation.

\item \textbf{Mismatched Delimiters.}  Parentheses or braces, when
  partially hidden or incorrectly aligned, triggered high local
  entropy.  In multiple examples, the text around an unmatched opening
  brace also included partially garbled subsequent tokens.

\item \textbf{Greek Letter Confusions.}  Letters like $\pi$ and
  $\rho$ can be mistaken for English `p` or `n`.  Such confusion led to
  upward spikes in entropy whenever the contextual cues about whether
  the symbol was a math variable or plain text were weak (e.g.\ a
  small snippet of math inline with normal prose).

\item \textbf{Symbol Dropping.}  Entire markers such as underscores
  (for subscripts) sometimes vanished at lower resolutions, again
  strongly increasing GPT--4o’s local uncertainty.  The missing
  symbol also tended to shift subsequent tokens, compounding the
  script confusion.

\end{enumerate}

Though not exhaustive, these examples underscore that the
sliding--window method successfully funnels attention toward typical
failure modes of math OCR.  In a more automated pipeline, it could
function as a triage mechanism for challenging pages.

\subsection{Annotator Effort}\label{subsec:reviewer_effort}

Our human annotators reported that spotting error-prone tokens became
faster when the three highest-entropy windows were highlighted in
color.  Annotators repeatedly skipped those portions of the text with
low entropy, focusing their scrutiny on only 10--15\,\% of the tokens.
While we did not attempt formal time--tracking, multiple participants
commented that having a “quick visual highlight” prevented them from
manually rereading entire pages for minor mistakes.  Therefore, even
without advanced metrics, the hotspots appear to reduce the overhead
of verifying GPT’s output.

The main frustration noted was that \emph{some} windows were “false
alarms,” in the sense that no clear error was discovered there.  These
were often short sequences of punctuation near math expressions; the
model seemingly was uncertain but did not produce an actual error.
Targeted re-prompting (Section~\ref{subsec:re_prompt}) might
resolve these borderline cases.

\subsection{Observations on Window Size and Resolution}
\label{subsec:observations}

Finally, we examined how the window length $W$ and the input resolution
interact in producing hotspots:

\paragraph{Window Length.}
In a pilot trial with $W=1$, numerous isolated tokens scored high
entropy, but the approach sometimes fragmented a single error across
several micro--windows.  At the other extreme, a large window ($W=20$)
could blur distinct issues together (e.g.\ a missing brace in one
equation and a Greek letter error in another).  Our final choice,
$W=10$, appeared to strike a suitable balance, grouping related tokens
without spanning too many lines of text.

\paragraph{Resolution.}
Unsurprisingly, the 72\,dpi transcripts included more frequent
hotspots, reflecting the model’s overall difficulty in interpreting
blurry details.  Meanwhile, for images at 300\,dpi, the number of
high-entropy windows typically dropped by 80\% or more.

In short, the findings confirm that our sliding-window entropy
heatmaps highlight local error-prone sections of GPT-based OCR output.
Although some caution is warranted around false alarms and parameter
choices like~$W$, this approach holds promise for faster, more focused
review of machine-transcribed mathematical content.
\section{Discussion}
\label{sec:discussion}

Our proof-of-concept study shows that \emph{sliding-window entropy} can direct human reviewers toward local regions of GPT--based OCR output where recognition errors are most likely to occur.  In contrast to prior, \emph{global} entropy approaches that yield a single scalar per page, our method yields a fine-grained “heat map” over the output tokens, identifying sub-sequences (hotspots) with heightened uncertainty.

\subsection{Why Local Entropy Peaks Reflect Actual Error Zones}
At a high level, an elevated local entropy arises when the model’s probability mass is spread over multiple competing tokens.  Such competition frequently stems from ambiguous visual cues (e.g.\ partial subscript, unclear symbol boundaries) or from parsing difficulties (e.g.\ a suspicious brace or bracket).  The data indicate that these same ambiguous visual features are precisely where practical OCR mistakes occur: misread minus signs, mismatched parentheses, or dropped subscripts.  Consequently, windowed entropy acts as a simple proxy for the model’s internal hesitation—highlighting precisely those tokens that “look suspicious” to GPT--4o.

Furthermore, our qualitative review suggests that, even if the system does not always produce a \emph{devastating} error at each high-entropy position, the suspicious region often contains subtle misalignments or unusual tokens that warrant a closer human look.  This is especially relevant in mathematical documents, where even minor symbol confusion can derail an entire expression.

\subsection{Integration into Digitisation Pipelines}
In practical workflows (e.g.\ arXiv paper ingestion or scientific scanning in digital libraries), volume is large and manual proofreading resources are limited.  Our results show that focusing reviewer effort on a handful of high-entropy windows can reduce the necessity of line-by-line scanning.  Implemented as a postprocessing step, the proposed module could:

\begin{enumerate}
\item \textbf{Read Output Logprobs:} Extract truncated probabilities from the GPT model as it decodes each text token.
\item \textbf{Compute Windowed Entropies:} Obtain $A_i$ for each sub-sequence of length $W$ (Sections~\ref{subsec:sliding_window_aggregation}--\ref{subsec:hotspots}).
\item \textbf{Highlight or Re-Prompt:} Mark the top-$M$ windows in the final transcript or pass them back to GPT--4o for a localized re-check.
\end{enumerate}

Notably, the entire process is straightforward to script, requiring $O(n)$ time once per-token log-probabilities are available.  This overhead seems negligible compared to the time saved in human correction.

\subsection{Limitations of Qualitative Conclusions}
Because our experiments relied mostly on human inspection of a small sample of pages, we acknowledge that the present results do not yield rigorous quantitative benchmarks, such as recall or precision rates for spotting errors.  Instead, we have offered a “first evidence” that local entropy spikes do align with the problematic tokens in practice.  Further expansions of the dataset, or a more systematic labeling of errors at scale, could confirm these findings under broader conditions.

Additionally, we have not extensively dissected \emph{false-positive hotspots} (entropy spikes that do \emph{not} contain real errors).  In the handful of such windows we observed, the model was apparently uncertain but still arrived at correct symbols.  Tailoring the method to further reduce these spurious alerts might require additional heuristics or re-prompting strategies.

\subsection{Broader Implications}
Despite these limitations, our findings resonate with the broader trend of integrating fine-grained uncertainty measures into large language model pipelines.  By focusing on local pockets of doubt, end users can apply targeted verification or domain-specific knowledge, boosting both efficiency and trustworthiness.  We see parallels in other multimodal tasks: for instance, scanning a neural speech recognizer’s token-level entropies can likewise flag unclear utterances for a second pass.  Extending such local entropy maps to more advanced tasks—like multi-column page layouts, embedded figures, or even interactive editing—presents exciting directions for future work.

In conclusion, these observations reinforce the general principle that “where the model is least certain” often overlaps with “where real OCR errors lurk.”  The sliding-window approach, while simple, appears surprisingly effective as a lightweight mechanism to triage GPT-based OCR outputs for potential mistakes.

\section{Conclusion and Future Work}\label{sec:conclusion_and_future_work}

In this paper, we presented a novel proof-of-concept framework for localizing OCR errors in GPT-based mathematical transcripts using per-token entropy signals. By converting per-token log-probabilities into a practical “uncertainty landscape,” our method pinpointed short contiguous segments—hotspots—most likely to contain recognition problems. The sliding-window averaging approach proved to be a lightweight and interpretable means of detecting local failures, as demonstrated on a small-scale corpus of scanned research pages rendered at several resolutions.

Qualitative evaluation showed that high-entropy windows align with various OCR pitfalls, including missing symbols, mismatched delimiters, and confusions among visually similar characters. Although our findings relied on human annotation over a modest set of 36 image samples, the strong anecdotal evidence suggests that this uncertainty-driven technique can help researchers or proofreaders focus attention on precisely those sub-sequences that merit closer review. Notably, it operates with minimal overhead, given that most GPT-based APIs already expose top-k token probabilities.

Nonetheless, we acknowledge certain limitations. First, the overall study size remains small, so additional experiments with larger and more diverse datasets are necessary to confirm the generality of our insights. Second, our ‘‘hotspot’’ detection occasionally flagged high-entropy regions that did not contain outright errors, pointing to a need for improved thresholds or follow-up diagnostics (e.g., automated re-prompting). Finally, the method was tested primarily on text with stationary mathematical expressions—tables, complex layouts, or heavily multi-column documents might require more advanced segmentation and tracking techniques.

Looking ahead, several extensions and directions emerge:
\begin{itemize}
\item \textbf{Automated Re-Prompting.} The proposed hotspot detection can be combined with a selective re-querying of GPT, requesting localized corrections only for suspicious subsequences.
\item \textbf{Fine-Grained Error Classification.} Beyond marking high-uncertainty tokens as potential errors, a downstream classifier could distinguish among error types (e.g., missing symbol vs.\ symbol confusion) to inform specialized correction modules.
\item \textbf{Adaptive Windowing.} Varying the window length automatically in response to local text features (such as formula boundaries) could enhance precision in highlighting exactly the relevant tokens.
\item \textbf{Multi-Task and Multilingual Extensions.} The same sliding-window entropy approach may prove useful in other scenarios (e.g., handwriting recognition, multilingual OCR) where local token competition signals confusion.
\item \textbf{Larger-Scale Benchmarking.} A more expansive and systematic data collection—covering more diverse page layouts, fonts, and scanning artifacts—would facilitate rigorous quantitative comparisons with alternative error-spotting methods.
\end{itemize}

In conclusion, our entropy heat-mapping framework illustrates how local confidence signals can offer a powerful lens into GPT-based OCR pipelines. By exposing precisely which tokens are most uncertain, it sets the stage for deeper refinements and targeted corrections, thereby improving the reliability of OCR outputs while minimizing human proofreading workloads. We hope that this proof-of-concept fosters further research on token-level uncertainty in large language models and drives the development of more reliable, transparent multimodal transcription systems.



\section*{Declarations}

\subsection*{Conflict of Interest}
The authors have no relevant financial or non-financial interests to disclose. The authors have no conflict of interest to declare for the content of this article.

\subsection*{Code Availability}
The Python code used in this study, developed by the authors, is openly accessible on GitHub at
\url{https://github.com/Alexei-WLU/scan2latex-entropy}. This repository contains all scripts necessary to reproduce the experimental results presented in this paper.


\end{document}